%% file: DSM.tex
\documentclass[twoside,11pt]{article}

\usepackage{jmlr2e}
\usepackage{subfig}
\usepackage{float}
\usepackage{lineno}
\usepackage[colorlinks=false]{hyperref}
\usepackage{natbib}
\usepackage{amsmath}
\usepackage{cleveref}

\ShortHeadings{Nonparametric Bayes dynamic modeling of relational data}{}
\firstpageno{1}

\begin{document}

\title{Nonparametric Bayes dynamic modeling of relational data}

\author{\name Daniele Durante \email durante@stat.unipd.it\\
      \addr Department of Statistical Sciences\\
       University of Padua\\
       Padua, Italy
\AND
       \name David B. Dunson \email dunson@duke.edu \\
       \addr Department of Statistical Science\\
     Duke University\\
       Durham, NC 27708-0251, USA
}

\editor{}

\maketitle
\linenumbers

\begin{abstract}%
 Symmetric binary matrices representing relations among entities are commonly collected in many areas.  Our focus is on dynamically evolving binary relational matrices, with interest being in inference on the relationship structure and prediction.  We propose a nonparametric Bayesian dynamic model, which reduces dimensionality in characterizing the binary matrix through a lower-dimensional latent space representation, with the latent coordinates evolving in continuous time via Gaussian processes.  By using a logistic mapping function from the probability matrix space to the latent relational space, we obtain a flexible and computational tractable formulation.  Employing P\'olya-Gamma data augmentation, an efficient Gibbs sampler is developed for posterior computation, 
with the dimension of the latent space automatically inferred.  We provide some theoretical results on flexibility of the model, and illustrate performance via simulation experiments.  We also consider an application to co-movements in world financial markets.
\end{abstract}
\section{Introduction}

\begin{figure}[t]
\centering
\includegraphics[height=6.5cm, width=15cm]{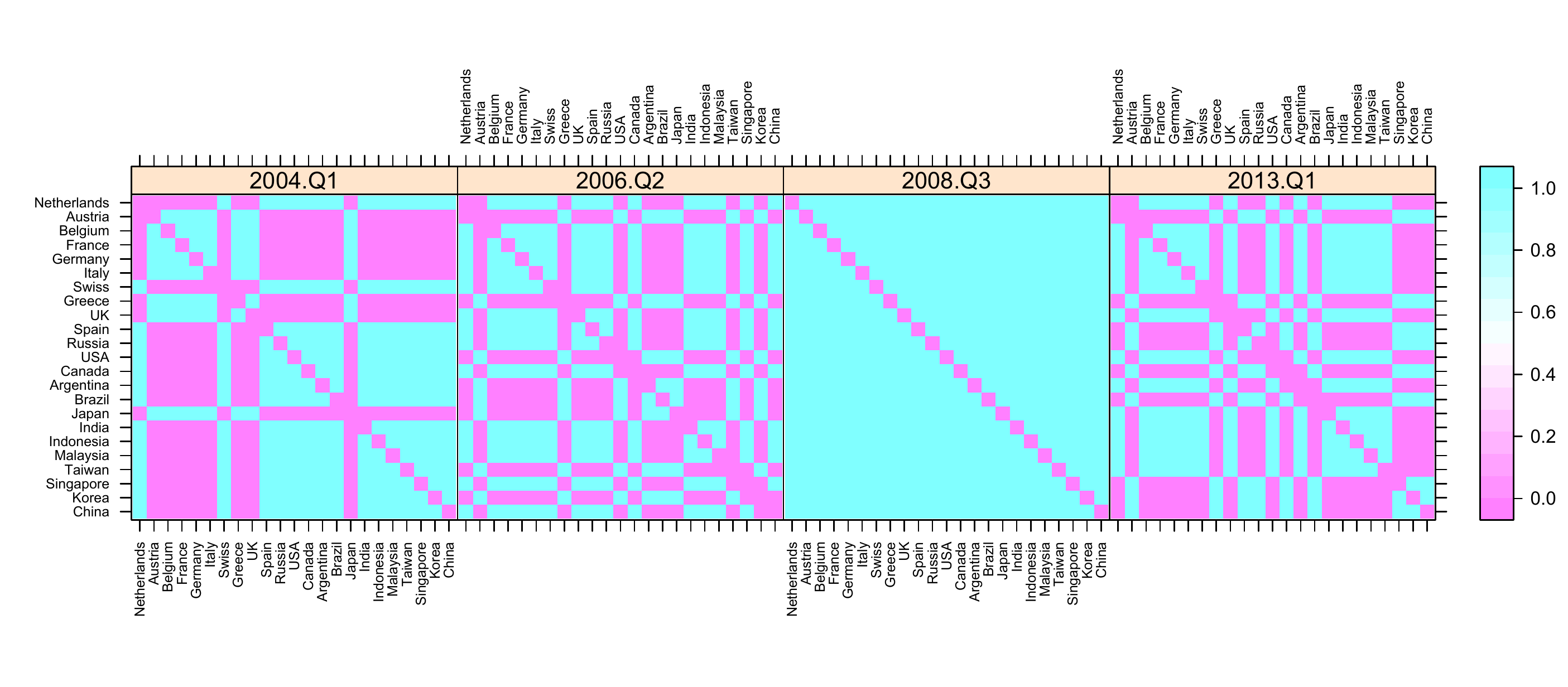}
\caption{\footnotesize{Dynamic co-movements in world financial markets.}}
\label{F1}
\end{figure}
Relational data often take the form of a symmetric binary matrix, with entries indicating the presence or absence of links between pairs of individuals or entities.  In dynamic settings, the links and the set of entities under consideration can change over time, and interest focuses on inferences on the time varying relational structure and in prediction.  Examples include social network analysis, in which links encode friendship networks among individuals, and broader relational settings in which closeness between a pair of units (products, stimuli, countries, companies, etc) is expressed on a binary scale.  Figure \ref{F1} shows an example of time-varying binary similarity matrices encoding dynamic co-movements in National Stock Market Indices from 2004 to 2013. Co-movements among a set of assets or market indices are typically analyzed via time-varying covariance or correlation matrices of their corresponding log-returns $Z_t=[z_{1,t},z_{2,t},...,z_{V,t}]'$, $t=1,...,T$, (see e.g. \citealp{Tsay:2005}, \citealp{Wi:2010}, \citealp{Naka:2012}, \citealp{Dur:2013}); we instead provide a different and not yet fully explored direction of research by treating co-movements as dynamic relational data, shifting our attention from $Z_t$ to the $V \times V$ time-varying symmetric matrices $\{Y_t, \ t \in \mathcal{T} \subset \Re^{+}\}$. The matrix $Y_t$ has entries $y_{ij,t}=y_{ji,t}=1$ if index $i$ and index $j$ move in the same direction at time $t$ (i.e. $z_{i,t}>0$ and $z_{j,t}>0$, or $z_{i,t}<0$ and $z_{j,t}<0$)  and $y_{ij,t}=y_{ji,t}=0$ if they move in opposite directions (i.e. $z_{i,t}>0$ and $z_{j,t}<0$, or $z_{i,t}<0$ and $z_{j,t}>0$). Co-movements indicate similarity in the indices.  

A rich literature is available on modeling similarity or dissimilarity matrices, with Multidimensional Scaling (MDS) providing a widely used technique for graphically representing units in a Euclidean space conditionally on their pairwise dissimilarity measures. General theory and applications are available for Euclidean distances and rank dissimilarities (see 
\citealp{Cox:2001}), with subsequent developments in a Bayesian framework (\citealp{Oh:2001}, \citealp{Oh:2007}) improving the overall performance, but subject to possible issues due to non-identifiable latent coordinates, lack of full conditional conjugacy
and absence of an automatic procedure for learning the dimension of the latent space. Moreover, generalizations in the dynamic case are lacking, with only few recent proposals restricted to specific applications for discrete time evolution \citep{Jam:2012}.  

When binary similarity or dissimilarity matrices are analyzed, the previous procedures prove to be inappropriate or impractical \citep{Hol:1982}, with predicted values outside the probability range and a large number of tied ranks for each unit in non-metric MDS applications. Spatial analysis of choice data (\citealp{De:1987}, \citealp{De:1999}) provides a possible generalization of MDS for binary variables, with recently developed algorithms available also in the dynamic case  \citep{Sark:2007}. However, questionable independence assumptions are required to ease maximum likelihood estimation, and Bayesian extensions  \citep{De:1999} to overcome this problem lack scalability in selecting the dimensionality of the latent space via cross-validation methods. Moreover, dynamic extensions via the Kalman filter rely on first and second order Taylor expansions for the observation model, providing difficulties in the derivation of theoretical properties for the exact formulation and requiring a sufficient number of observations to meet the Gaussian assumption. These models are specifically tailored for embedding problems in 2-mode co-occurrence data recording links between two different types of entities (i.e. consumer-products, author-words). Our focus is instead on dynamic modeling for one-mode binary matrices. 

There is a growing body of literature in social networks on model-based statistical analysis of one-mode binary matrices, traditionally focusing on overly-restrictive models, such as \citet{Erd:1959}, the $p_1$ model \citep{Hol:1981} and the Exponential Random Graph Model (ERGM) \citep{Fra:1986}, with generalizations for dynamic inference available via discrete temporal ERGM \citep{Rob:2001} and hidden temporal ERGM \citep{Guo:2007}. ERGMs have had growing popularity, but have a number of drawbacks.  Estimation relies on pseudo-likelihood \citep{Stra:1999} and approximate MCMC methods \citep{Sni:2002}, due to the computational intractability in a fully likelihood approach.  Solutions can be degenerate or nearly-degenerate \citep{Han:2003}, and questions remain about coherence, inflexibility and other key issues.

An alternative class of models focus on clustering the nodes, based on the pattern of inter-connections in the network. Stochastic Block Models (SBM) \citep{Now:2001} provide a common framework, with the Infinite Relational Model (IRM) \citep{Kem:2006} allowing an unknown number of clusters via a Dirichlet process.  Dynamic SBMs have been recently considered (\citealp{Ish:2010},\citealp{Ya:2011}, \citealp{Xu:2013}). \citet{Ish:2010} focus on discrete dynamic evolution via a hidden Markov model. \citet{Xu:2013} accommodate continuous time analysis via a state space formulation, but require sufficient numbers of observations in each block to meet Gaussian assumptions for the sample mean.  They use the extended Kalman filter to linearize the observation equation, leading to questions of accuracy.

We dynamically model binary relational matrices by embedding the nodes in a low-dimensional latent Euclidean space, with coordinates evolving in continuous time via Gaussian processes and edge probabilities constructed utilizing a logistic mapping function from the probability matrix space to the dot product of the latent coordinates. Hence, we are most closely related to the literature on latent space models \citep{Hoff:2002} and Mixed Membership Stochastic Block models (MMSB) \citep{Ai:2008}, which allow each node to belong to multiple blocks with fractional membership.  Dynamic latent space models \citep{Sark:2005} and MMSB models \citep{Xi:2010} incorporate Gaussian perturbations in discrete time and state space models, respectively.  Posterior computation relies on several layers of approximation without theory available to justify accuracy.  In contrast, we provide a simple Gibbs sampling algorithm for our model, which converges to the exact posterior and infers the dimension of the latent space automatically.

The paper is organized as follows. In Section 2, we describe the general model structure with particular attention to prior specification and theoretical properties. Section 3 provides the Gibbs sampling steps. A simulation study is examined in Section 4, and an application to quarterly co-movements in world financial markets is presented in Section 5.

\section{Dynamic Latent Space Model}
\subsection{Notation and Motivation}
Let $Y_t$ be the symmetric binary similarity matrix at time $t \in \mathcal{T}$ and $\pi(t)$ be the corresponding symmetric probability matrix having entries $\pi_{ij}(t)=\pi_{ji}(t)=\mbox{pr}(y_{ij,t}=1)$ for every $i=1,\dots,V$ and $j=1,\dots,V$. Letting
\begin{eqnarray}
y_{ij,t} | \pi_{ij}(t) \sim \mbox{Bern}(\pi_{ij}(t)),
\label{eq1}
\end{eqnarray}
independently for each $i=2,\ldots,V$ and $j=1,\ldots,i-1$, our aim is to define a prior $\Pi_{\pi}$ for the collection of time-varying probability matrices $\pi_{\mathcal{T}}=\{\pi(t), t \in \mathcal{T} \}$ with the goals being to (i) obtain a provably flexible specification, (ii) maintain simple computations, (iii) perform dimensionality reduction in order to scale to moderately large $V$ settings, (iv) allow unequal spacing and missing observations and (v) allow predictions including a measure of predictive uncertainty. Since the matrices are symmetric and the similarities or dissimilarities of a unit with itself are meaningless, we will focus on modeling the lower triangular part without taking into account the diagonal elements.

\subsection{Latent space dynamic model formulation}
We construct $\pi_{ij}(t)$ via a monotonic increasing link function $g(\cdot): \Re \rightarrow [0,1]$ mapping a latent similarity measure among units $i$ and $j$ at time $t$, $s_{ij}(t) \in \Re$, into the probability space. Specifically, we choose $g(\cdot)$ to be the distribution function of the logistic random variable, obtaining
\begin{eqnarray}
\mbox{E}[y_{ij,t}|\pi_{ij}(t)]=\pi_{ij}(t)=\frac{1}{1+e^{-s_{ij}(t)}}, 
\label{eq2}
\end{eqnarray}
for $i=2,\ldots,V$, $j=1,\ldots,i-1,$ and $t \in \mathcal{T}$. Without further assumptions on $s_{ij}(t)$, one needs to model separately $\frac{1}{2} V(V-1)$ stochastic processes, one for each time-varying similarity measure $s_{ij}(t)$, with $i=2,...,V$, $j=1,...,i-1$ and $t \in \mathcal{T}$, leading to burdensome computations as $V$ increases and failing to borrow information exploiting the underlying process inducing similarities among the units.   In order to reduce the dimensionality of the problem and to learn also the network structure among the units for every $t$, we express the similarity measures $s_{ij}(t)$ as a quadratic combination of a set of latent coordinates for unit $i$ and unit $j$. Specifically
\begin{eqnarray}
s_{ij}(t)=\mu(t)+x_{i}(t)'x_{j}(t),
\label{eq3}
\end{eqnarray}
where $x_{i}(t)=[x_{i1}(t), \ldots, x_{iH}(t)]'$ for $i=2,\ldots,V$ and $x_{j}(t)=[x_{j1}(t),\ldots,x_{jH}(t)]'$ for $j=1,\ldots,i-1$, are the vectors of latent coordinates of unit $i$ and $j$ respectively, giving rise, together with the baseline $\mu(t)$, to the similarity measure between the two units via a projection approach. According to this specification, units with latent coordinates in the same directions will have a higher probability of being similar (i.e. $y_{ij,t}=1$), while units with opposite coordinates are more likely to be dissimilar (i.e. $y_{ij,t}=0$). 

This formulation is also intuitive in practical applications. Recall our motivating example of finance, and assume for simplicity $\mu(t)=0$ and only two latent coordinates representing for example unexpected inflation and industrial production, respectively. Then indices of countries with features in the same directions will have a higher probability of co-moving, while countries with opposite unexpected inflation and industrial production will more likely move on different directions.

In matrix notation, equation (\ref{eq3}) can be rewritten as
\begin{eqnarray}
S(t)=\mu(t) 1_V1_V'+X(t)X(t)',
\label{eq3matr}
\end{eqnarray}
where $S(t)$ is a $V \times V$ real symmetric matrix with latent similarity entries $s_{ij}(t)$ and $X(t)=[x_{1}(t), x_{2}(t),\dots,x_{V}(t)]'$. Note that, assuming without loss of generality $\mu(t)=0$, the above decomposition is not unique. For example if we define $X(t)^{*}=X(t) Q$ with $Q$ a $H \times H$ orthogonal matrix, then $X(t)^{*}X(t)^{*'}=X(t) QQ'X(t)'=X(t)X(t)'$. If one is interested also in making inference on the latent coordinates matrix $X(t)$, different proposals are available in latent factor modeling to ensure identifiability via restrictions (see e.g. \citealp{Bo:1989}) or Procrustean transformations \citep{Hoff:2002}. However since our focus is on making inference and prediction on the probability matrices, we follow \citet{Gosh:2009} in avoiding identifiability constraints, as such constraints are not necessary to ensure identifiability of the induced similarity matrix $S(t)$. 

It is important to characterize the class of $\pi(t)$ matrices whose lower triangular elements can be represented as in (\ref{eq2}) with latent similarities decomposed as in (\ref{eq3matr}). Theorem \ref{teo1} and the corresponding Corollary \ref{cor1} state that for $H$ sufficiently large, the lower triangular matrix elements of any symmetric probability matrix have such a representation. For $H \geq V$, $\mathcal{X}_{X} $ denotes the space of all $V \times H$ dimensional matrices of arbitrary coordinate functions mapping from $\mathcal{X} \rightarrow \Re$ and $\mathcal{X}_{\mu}$ the space of all baseline mean functions. 

\begin{theorem}
Given a symmetric real matrix $S(t)$, $\forall \ t \in \mathcal{T}$, there exist $\{ X(t), \mu(t)\} \in \mathcal{X}_{X} \otimes \mathcal{X}_{\mu} $ such that $$ S(t)=\mu(t)\times 1_V1_V'+X(t)X(t)', \quad \forall \ t \in \mathcal{T}$$
\vspace{-10pt}
\label{teo1}
\end{theorem}
{\bf Proof}. Assume without loss of generality that $\mu(t)=0$ and take $H \geq V$. Consider
\[
X(t)=[ \ P(t)\Lambda(t)^{1/2} \ \ \ 0_{V\times (H-V)} \ ],
\]
where $P(t)$ is the matrix of the eigenvectors of $S(t)$ and $\Lambda(t)$ the diagonal matrix with the corresponding eigenvalues. Then $S(t)=P(t)\Lambda(t)P(t)' =X(t)X(t)'$, for every $t \in  \mathcal{T}$.

\begin{corollary}
Given a symmetric probability matrix $\pi(t)$, $\forall \ t \in \mathcal{T}$, there exist $\{ X(t), \mu(t)\} \in \mathcal{X}_{X} \otimes \mathcal{X}_{\mu} $ such that $$ \pi_{ij}(t)=\frac{1}{1+e^{-\mu(t)-\sum_{h=1}^{H} x_{ih}(t)x_{jh}(t)}}, \quad \forall \ t \in \mathcal{T}, \ i=2,...,V, \ j=1,...,i-1$$
\label{cor1}
\end{corollary}
{\bf Proof}. The proof follows immediately from Theorem \ref{teo1} and from the fact that the mapping from $s_{ij}(t)$ to $\pi_{ij}(t)$ is a one-to-one continuous increasing function.

\vspace{5pt}
This ensures that our specification is sufficiently flexible to characterize any true generating process, and hence can be viewed as nonparametric given sufficiently flexible priors for the components.

\subsection{Prior Specification}
We aim to specify independent prior distributions $\Pi_{X}$ and $\Pi_{\mu}$ for $X_{\mathcal{T}}=\{X(t), t \in \mathcal{T} \}$ and $\mu_{\mathcal{T}}=\{\mu(t), t \in \mathcal{T} \}$ in order to induce a prior $\Pi_{\pi}$ for $\pi_{\mathcal{T}}=\{\pi(t), t \in \mathcal{T} \}$ through (\ref{eq2}) and (\ref{eq3}).  This prior is carefully defined to have large support, favor simple and efficient computation, allow missing values, induce a continuous time specification, and allow learning of the latent space dimension.  \citet{Bhat:2011} proposed a useful approach for Bayesian learning of the number of latent factors in a model for a single large covariance matrix, and we extend their approach from independent Gaussian latent factors to Gaussian process latent factors. In particular, we let 
\begin{eqnarray*}
x_{ih}(\cdot) \sim \mbox{GP}(0,\tau_{h}^{-1}c_{X}),
\label{eq4}
\end{eqnarray*}
independently for all $i=1,...,V$ and $h=1,\dots,H$, with $c_{X}$ a squared exponential correlation function  $c_{X}(t,t')=\exp(-\kappa_{X} ||t-t' ||^{2}_{2})$, which allows for continuous time analysis and unequal spacing, and $\tau_{h}^{-1}$ a shrinkage parameter defined as  
\begin{eqnarray*}
\tau_{h}=\prod_{k=1}^{h}\vartheta_{k},\ \vartheta_{1}\sim \mbox{Ga}(a_{1},1), \ \vartheta_{k}\sim\mbox{ Ga}(a_{2},1), k \geq2.
\label{eq:5}
\end{eqnarray*}
Note that if $a_{2}>1$ the expected value for $\vartheta_{k}$ is greater than $1$. As a result, as $h$ goes to infinity, $\tau_{h}$ tends to infinity, shrinking $x_{ih}(\cdot)$, for every $i=1,\ldots,V$ towards zero. This leads to a flexible prior for $x_{ih}(\cdot) $ with a local shrinkage parameter $\tau_{h}^{-1}$ that favors many stochastic processes of latent coordinates being close to $0$ as $h$ increases. To conclude prior specification we choose
\begin{eqnarray*}
\mu(\cdot) \sim \mbox{GP}(0,c_{\mu}),
\label{eq5}
\end{eqnarray*}
with $c_{\mu}(t,t')=\exp(-\kappa_{\mu} ||t-t' ||^{2}_{2})$.

Before proceeding with posterior computation, we focus on the support of the induced prior $\Pi_{\pi}$ based on priors $\Pi_{X}$ and $\Pi_{\mu}$. Specifically we are interested in proving whether the prior can generate a time-varying symmetric probability matrix that is arbitrarily close to any function $\{\pi(t), t \in \mathcal{T} \}$. Intuitively, large support on  continuous symmetric similarity matrix functions $\{S(t), t \in \mathcal{T} \}$ relies on the continuity of the Gaussian process coordinate functions. Since for each fixed $t=t_0$, $x_{ih}(t_0)$ are independently Gaussian distributed, $X(t_0)X(t_0)'$ is distributed according to a sum of independent Wishart random variables. Combining the large support of the Wishart distribution with the one of the Gaussian for the baseline $\mu(t_0)$, provides large support for the induced prior $\Pi_{S}$. Since $\pi(t)$ is obtained via a one to one continuous increasing function of $S(t)$, we will map non-null probability subsets of the space of $S(t)$ into non-null probability subsets of the space of $\pi(t)$, providing the desired large support for the induced prior $\Pi_{\pi}$. Theorem \ref{teo2} states the large support property for $\Pi_S$, while Corollary \ref{cor2} provides the same property for $\Pi_\pi$ by combining results in the previous Theorem with the fact that $\pi(t_0)$ is defined as a monotonic increasing continuous function of $S(t_0)$. Proof of Theorem \ref{teo2} is provided in Appendix.

\begin{theorem}
Let $\Pi_{S}$ denote the induced prior on $\{S(t), t \in \mathcal{T} \}$ based on the specified prior $\Pi_{X}\otimes \Pi_{\mu}$ on $ \mathcal{X}_{X} \otimes \mathcal{X}_{\mu}$. Assuming $\mathcal{T}$ compact, for all continuous $S^*(t)$ and for all $\epsilon >0$
$$ \Pi_{S} \left( \sup_{t \in \mathcal{T}}||S(t)-S^*(t) ||_2<\epsilon \right)>0. $$
\label{teo2}
\end{theorem}
\begin{corollary}
Let $\Pi_{\pi}$ denote the induced prior on $\{\pi(t), t \in \mathcal{T} \}$ based on the specified prior $\Pi_{X}\otimes \Pi_{\mu}$ on $ \mathcal{X}_{X} \otimes \mathcal{X}_{\mu}$. Assuming $\mathcal{T}$ compact, for all continuous $\pi^*(t)$ and for all $\delta >0$
$$ \Pi_{\pi} \left( \sup_{t \in \mathcal{T}}||\pi(t)-\pi^*(t) ||_2<\delta \right)>0. $$
\label{cor2}
\end{corollary}
{\bf Proof}. Since the elements of $\pi(t)$ are defined as a one to one continuous mapping of the elements of $S(t)$ through the function $g(\cdot)$, by definition of continuity we have that for every $\delta>0$ there exists an $\epsilon>0$ such that
$$ \sup_{t \in \mathcal{T}}||g\left(S(t)\right)-g\left(S^*(t)\right) ||_2= \sup_{t \in \mathcal{T}}||\pi(t)-\pi^*(t) ||_2<\delta $$ 
for all $S(t)$ such that  $ \sup_{t \in \mathcal{T}}||S(t)-S^*(t) ||_2<\epsilon $, where $g\left(S(t)\right)$ means that the function $g(\cdot)$ is applied to every element of $S(t)$. Finally, since by Theorem \ref{teo2} the event $ \sup_{t \in \mathcal{T}}||S(t)-S^*(t) ||_2<\epsilon $ has non-null probability, it follows that the same holds for the event $\sup_{t \in \mathcal{T}}||\pi(t)-\pi^*(t) ||_2<\delta$, completing the proof.

\section{Posterior computation}
Posterior computation is performed adapting a recently proposed data-augmentation scheme based on a new class of P\'olya-Gamma distributions; for a detailed description see \citet{Po:2013}. The approach provides a strategy for fully Bayesian inference in models with binomial likelihoods, which bypasses the need for analytic approximations, while allowing us to exploit conjugacy for block updating.

The main result is that binomial likelihoods parameterized by log-odds can be represented as a mixture of Gaussians with respect to P\'olya-Gamma distributions. Specifically
\begin{eqnarray*}
\frac{(e^{\psi})^a}{(1+e^{\psi})^b}=2^{-b}e^{z\psi} \int_{0}^{+\infty}e^{-\omega \psi^2/2}p(\omega) \partial \omega,
\end{eqnarray*}
where $z=a-b/2$ and $\omega \sim \mbox{PG}(b,0)$, with $\mbox{PG}(b,c)$ denoting the P\'olya-Gamma random variable with parameters $c \in \Re$ and $b>0$. When $\psi=x' \beta$ is a linear predictor, and a Gaussian prior is considered for $\beta$, full conditional conjugacy is ensured for Bayesian inference on the coefficients. Moreover the implied conditional distribution for $\omega$, given $\psi$, is again P\'olya-Gamma, providing a simple Gibbs sampler alternating between two main steps. Specifically, letting $y_i$ be the number of successes and $x_i=[x_{i1},...,x_{ip}]'$ the vector of regressors for every observation $i=1,...,N$, and assuming a Bayesian logistic regression setting where $y_i \sim \mbox{Bern}(1/[1+e^{-\psi_i}])$, $\psi_i=x_i' \beta$ and $\beta$ having Gaussian prior $\beta \sim \mbox{N}_p(b,B)$, the Gibbs alternates between
\begin{eqnarray*}
\omega_i |\beta, x_i \sim \mbox{PG}(1,x_i' \beta)\quad \mbox{and} \quad \beta | y,\omega, x \sim \mbox{N}_p(\mu_{\beta}, \Sigma_{\beta}),
\end{eqnarray*}
where $\Sigma_{\beta}=(X' \Omega X+B^{-1})^{-1}$ and $\mu_{\beta}=\Sigma_{\beta}(X'z+B^{-1}b)$; with $z=[y_1-1/2,....,y_N-1/2]'$ and $\Omega$ is the diagonal matrix with $\omega_i$'s entries. 

Recalling model (\ref{eq1}), with probabilities defined as in (\ref{eq2}) and latent similarities from (\ref{eq3}), for $i=2,...,V$, $j=1,...,i-1$ and $t \in \mathcal{T}_{0}=\{t_1,...,t_{T} \}$, and taking a fixed truncation level $H^{*}$ for the number of latent coordinates, the Gibbs sampler for our model, is:
\begin{enumerate}
\item{Update each augmented data $\omega_{ij,t}$ from the full conditional P\'olya-Gamma posterior:
\begin{eqnarray*}
\omega_{ij,t}|x_{i}(t),x_{j}(t),\mu(t) \sim \mbox{PG}\left(1,\mu(t)+\sum_{h=1}^{H^{*}} x_{ih}(t)x_{jh}(t)\right)
\end{eqnarray*}
for every $i=2,...,V$, $j=1,...,i-1$ and $t \in \mathcal{T}_{0}=\{t_1,...,t_{T} \}$.}

\item{Given $\{y_{ij,t}\}$, $X(t)$ and $\{\omega_{ij,t}\}$, the P\'olya-Gamma data augmentation scheme ensures full conditional Gaussian posterior for $\mu(t)$ with $t \in \mathcal{T}_{0}=\{t_1,...,t_{T} \}$, of the form
{\footnotesize{
\begin{eqnarray*}
\left[ \begin{array}{c}
\mu(t_1)\\
\mu(t_2)\\
\vdots\\
\mu(t_T)
 \end{array} \right]| \{y_{ij,t}\}, X(t), \{\omega_{ij,t}\} \sim \mbox{N}_{T}\left(\Sigma_{\mu}\left[ \begin{array}{c}
\sum_{i=2}^{V}\sum_{j=1}^{i-1}(y_{ij,t_1}-0.5-\omega_{ij,t_1}x_{i}(t_1)'x_{j}(t_1))\\
\sum_{i=2}^{V}\sum_{j=1}^{i-1}(y_{ij,t_2}-0.5-\omega_{ij,t_2}x_{i}(t_2)'x_{j}(t_2))\\
\vdots\\
\sum_{i=2}^{V}\sum_{j=1}^{i-1}(y_{ij,t_T}-0.5-\omega_{ij,t_T}x_{i}(t_T)'x_{j}(t_T))
 \end{array} \right],\Sigma_{\mu}\right)
\end{eqnarray*}}}
With $\Sigma_{\mu}=\left[\mbox{diag}\left(\sum_{i=2}^{V}\sum_{j=1}^{i-1}\omega_{ij,t_1},\dots,\sum_{i=2}^{V}\sum_{j=1}^{i-1}\omega_{ij,t_T} \right)+ K_{\mu}^{-1}\right]^{-1}$, and $K_{\mu}$ the Gaussian process covariance matrix with $[K_{\mu}]_{ij}=\exp(-\kappa_{\mu} ||t_i-t_j ||^{2}_{2})$.}
\item{Update the time-varying latent coordinate vector $\{x_{v}(t)=[x_{v1}(t),...,x_{vH^*}(t) ]'\}_{t=t_1}^{t_T}$ for every unit $v=1,...,V$ from its conditional posterior. Specifically, conditionally on $X^{(-v)}=\{x_{j}(t): j \neq v, t \in \mathcal{T}_{0}=\{t_1,...,t_{T} \} \}$, $\mu=[\mu(t_1),\dots,\mu(t_T)]'$, $\{y_{ij,t}\}$, $\{\omega_{ij,t}\}$, $\{\tau_{h} \}$ and defining $y_{ij}=[y_{ij,t_1},\dots,y_{ij,t_T}]'$ and $\pi_{ij}=[\pi_{ij,t_1},\dots,\pi_{ij,t_T}]'$, let $Y^{(v)}$ be the vector obtained by stacking sub-vectors $y_{ij}$ for all the couples $(i,j)$ such that $i=v$ or $j=v$, with $i>j$; and $\pi^{(v)}$ the corresponding vector of probabilities, then
\begin{eqnarray}
\mbox{logit}(\pi^{(v)})| X^{(-v)}, \mu(t)&=&1_{V-1}\otimes \mu +\tilde{X}_{x_{v}(t)}\beta_{x_{v}(t)}
\label{eq:7}
\end{eqnarray}
where $\beta_{x_{v}(t)}=[x_{v1}(t_1),...,x_{v1}(t_T),x_{v2}(t_1),...,x_{v2}(t_T), \dots,x_{vH^*}(t_1),...,x_{vH^*}(t_T) ]'$ with prior, according to GP formulation, $\beta_{x_{v}(t)} \sim N_{T\times H^*}\left( 0,\mbox{diag}[\tau_{1}^{-1},...,\tau_{H^*}^{-1}] \otimes K_{x} \right)$ and $\tilde{X}_{x_{v}}(t)$ a matrix of regressors with entries suitably chosen from the elements of $X^{(-v)}$ in order to reproduce the equality:
$$\mbox{logit}(\pi_{ij}(t))| X(t), \mu(t)=\mu(t)+\sum_{h=1}^{H^*} x_{ih}(t)x_{jh}(t)$$
for all the probabilities $\pi_{ij}(t)$ such that $i=v$ or $j=v$, with $i>j$ and $ t \in \mathcal{T}_{0}=\{t_1,...,t_{T} \}$. Model (\ref{eq:7}) is a proper logistic regression with linear predictor, therefore, according to our P\'olya-Gamma sampling scheme, we update the vector of time-varying coordinates $\{x_{v}(t)=[x_{v1}(t),...,x_{vH^*}(t) ]'\}_{t=t_1}^{t_T}$, represented by $\beta_{x_{v}(t)}$ by sampling from:
\begin{eqnarray*}
\beta_{x_{v}(t)}|X^{(-v)},\{\mu(t)\},\{y_{ij,t}\},\{\omega_{ij,t}\},\{\tau_{h} \} \sim N_{T\times H^*}\left( \mu_{x_{v}(t)},\Sigma_{x_{v}(t)} \right)
\end{eqnarray*}
with
\begin{eqnarray*}
 \Sigma_{x_{v}(t)}&=&\left( \tilde{X}_{x_{v}(t)}' \Omega_{x_{v}(t)} \tilde{X}_{x_{v}(t)} +\mbox{diag}[\tau_{1},...,\tau_{H^*}] \otimes K_{x}^{-1}  \right)^{-1}\\
\mu_{x_{v}(t)}&=&\Sigma_{x_{v}(t)}\left[ \tilde{X}_{x_{v}(t)}' \left(Y^{(v)}-1_{V-1}\otimes1_{T}\times 0.5 -1_{V-1}\otimes  \mu\right)\right]
\end{eqnarray*}
and $\Omega_{x_{v}(t)}$ is the diagonal matrix with the corresponding P\'olya-Gamma augmented data.}

\item{Conditioned on $X(t)$ and $\{\tau_{h} \}$, sample the global shrinkage hyperparameters from
\begin{eqnarray*}
\vartheta_{1}|X(t), \tau^{(-1)} &\sim& \mbox{Ga} \left(a_{1}+\frac{V\times T \times H^{*}}{2},1+\frac{1}{2}\sum_{l=1}^{H^{*}}\tau_{l}^{(-1)}\sum_{i=1}^{V}x_{il}'K_{x}^{-1}x_{il}\right) \quad \quad \quad \quad \nonumber \\
\vartheta_{h}|X(t), \tau^{(-h)} &\sim& \mbox{Ga}\left(a_{2}+\frac{V\times T \times (H^{*}-h+1)}{2},1+\frac{1}{2}\sum_{l=1}^{H^{*}}\tau_{l}^{(-h)}\sum_{i=1}^{V}x_{il}'K_{x}^{-1}x_{il}\right)
\end{eqnarray*}
Where $\tau_{l}^{(-h)}=\prod_{t=1,t\neq h}^{l} \vartheta_{t}$ for $h=1,...,H^*$ and $x_{il}=[x_{il}(t_1), \dots, x_{il}(t_T)]'$.}
\end{enumerate}

We can easily handle missing values by adding a further step imputing the unobserved binary similarities from their conditional distribution given the current state of the chain. Specifically:
\begin{description}
\item{5. Given $X(t)$ and $\mu(t)$ sample each missing value from its conditional distribution
\begin{eqnarray*}
y_{ij,t}=y_{ji,t}|X(t),\mu(t) \sim \mbox{Bern}\left(\frac{1}{1+e^{-\mu(t)-\sum_{h=1}^{H^*} x_{ih}(t)x_{jh}(t)}} \right), \quad i>j.
\end{eqnarray*}}
\end{description}
Step 5 provides also a strategy for predicting new outcomes. Specifically, if we are interested in making inference on future $\pi(t_{T+1})$ with $t_{T+1}>t_{T}$ given the observed similarity matrices $Y_t$, $t \in \mathcal{T}_{0}=\{t_1,...,t_{T} \}$, then we can simply perform the previous posterior computations adding to the observed dataset $\{Y_t\}_{t\in \mathcal{T}_{0}}$ a new matrix $Y_{t_{T+1}}$ of missing values and make inference on the predictive posterior distribution using the samples of the Markov chain for  $\pi(t_{T+1})$.

\section{Simulation Study}
We provide a simulation study with the aim to evaluate the performance of the proposed model in analyzing a dataset constructed to mimic also a possible generating process in the finance application. The focus is on the ability to correctly reconstruct the true underlying processes, and also on the performance with respect to out of sample predictions. We also provide a comparison between our proposed approach and the estimated probability process for each time-varying binary outcome when using only temporal information without exploring matrix structure, showing graphically the sub-optimality of the latter in terms of efficiency and bias.
\begin{figure}[t]
\centering
\includegraphics[height=10cm, width=13cm]{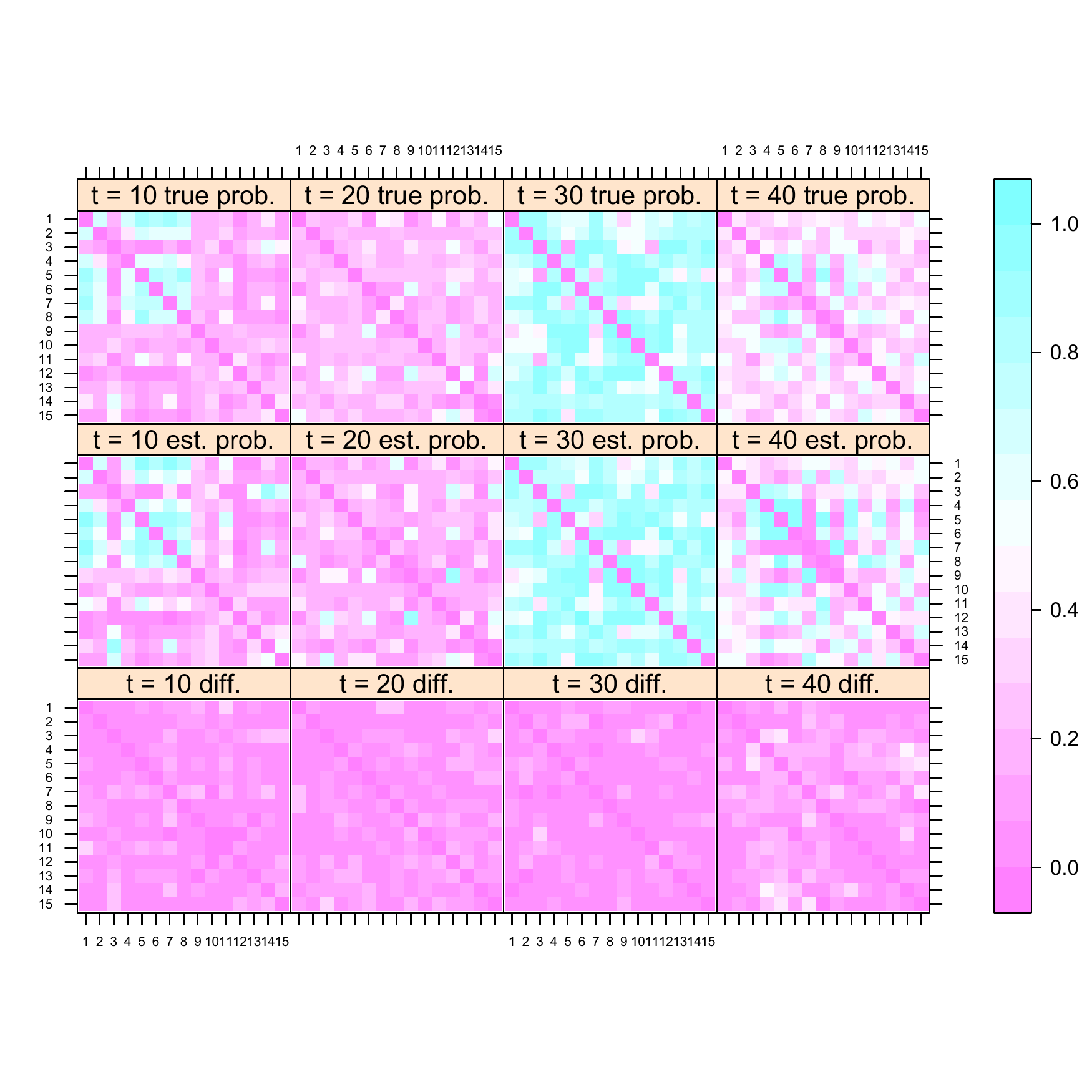}
\caption{\footnotesize{For some selected times $t$, plot of the true probability matrix $\pi(t)$ (top), posterior mean $\hat{\pi}(t)$ from our model (middle) and absolute value of the difference among the two $|\pi(t)-\hat{\pi}(t)|$ (bottom). }}
\label{F2}
\end{figure}

\subsection{Estimating Performance}
We generate a set of $15 \times 15$ time varying $Y_{t}$ matrices with $t$ in the discrete set $\mathcal{T}_0=\{1,2, \dots,40 \}$. Each $y_{ij,t}$ is simulated according to (\ref{eq1}) with probabilities obtained from (\ref{eq2}) and (\ref{eq3}), generating $\{\mu(t)\}_{t=1}^{40}$ from a $\mbox{GP}(0,c_{\mu})$ with length scale $\kappa_{\mu}=0.01$ and choosing $2$ time-varying latent coordinates $\{x_{i1}(t)\}_{t=1}^{40}$, $\{x_{i2}(t)\}_{t=1}^{40}$ from Gaussian processes with length scale $\kappa_{x}=0.01$, independently for each unit $i=1,...,15$. To evaluate the out of sample predictive performance we take $Y_{40}$ to be a matrix of missing values, and assume similarities between units $10$ and $11$ and all the others, missing at times $t=20,...,25$ to assess the behavior with respect to missing data. For inference we choose a truncation level $H^{*}=10$, length scales $\kappa_{\mu}=\kappa_{x}=0.05$ and set $a_1=a_2=2$ for the shrinkage parameters. We ran $5{,}000$ Gibbs iterations which proved to be enough for reaching convergence and discarded the first $1{,}000$. Mixing was assessed by analyzing the effective sample sizes of the MCMC chains for the quantities of interest (i.e. $\pi_{ij}(t)$, for $i=2,...,V$, $j=1,...,i-1$ and $t \in \mathcal{T}_0$) after burn-in. We found most of these values concentrating around $\approx 1{,}700$ effective samples on a total of $4{,}000$, providing a good mixing result.

The comparison in Figure \ref{F2} between true probability matrices and their corresponding posterior mean for some selected time $t$, highlights the good performance of our approach in correctly estimating the true latent process and making predictions. The latter can be noticed by comparing true and estimated probability matrices at $t=40$, recalling that in our simulation we assumed $Y_{40}$ having missing entries and we were interested in analyzing the predictive performance of our model with respect to $\pi(40)$. Similar results are provided by the plot of true $\pi_{ij}(t)$ against the corresponding estimates $\hat{\pi}_{ij}(t)$ and by the ROC curve in Figure \ref{F3} having an area underneath of $0.87$.

\begin{figure}[t]
\centering
\includegraphics[height=5.5cm, width=12cm]{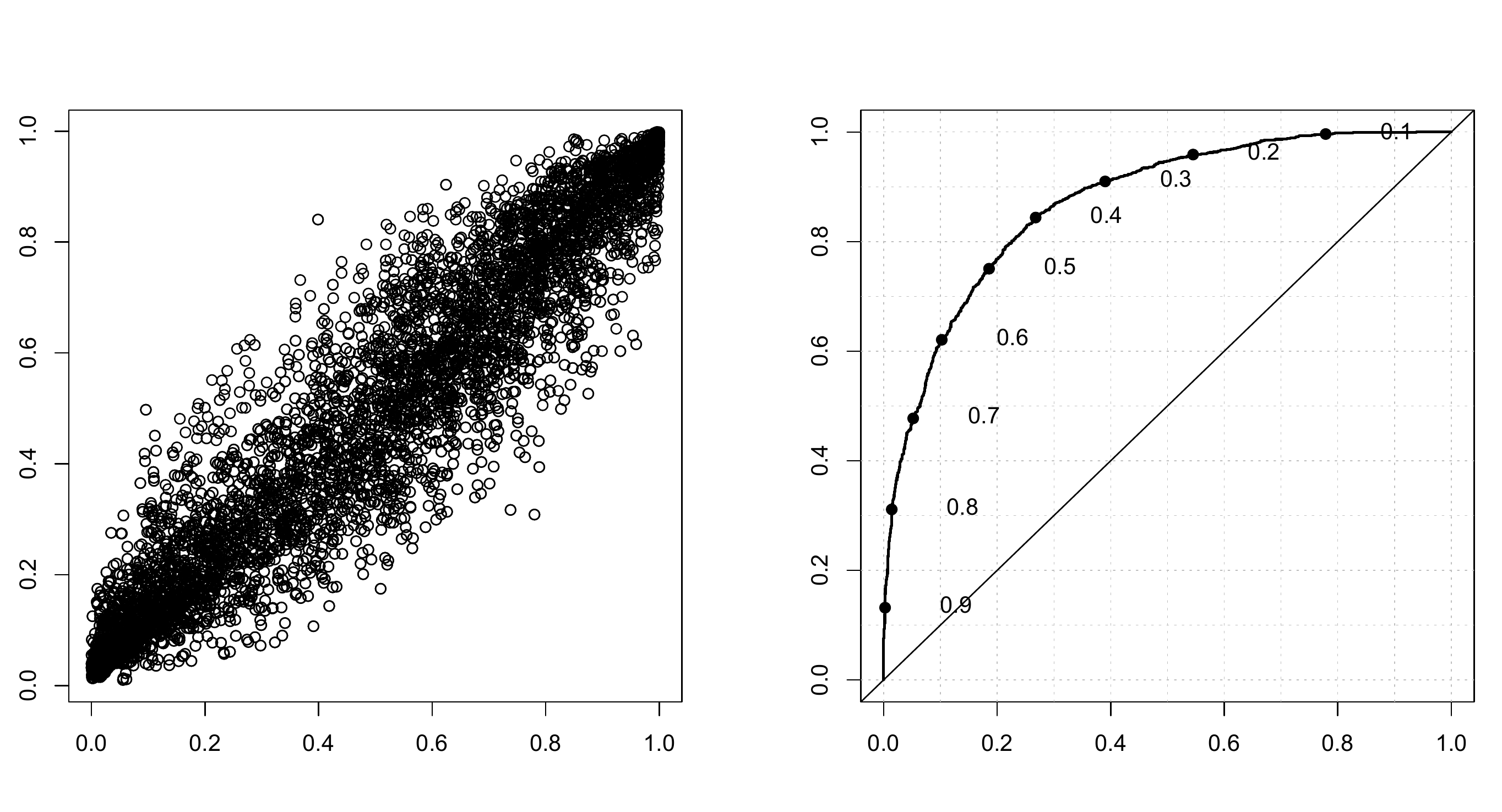}
\put (-370,80) {$\hat{\pi}_{ij}(t)$}
\put (-270,-5) {$\pi_{ij}(t)$}
\caption{\footnotesize{Left: plot of true probabilities $\pi_{ij}(t)$ versus their corresponding posterior mean $\hat{\pi}_{ij}(t)$, for $i=2,...,V$, $j=1,...,i-1$ and $t \in \mathcal{T}_0$. Right: ROC curve generated using $\hat{\pi}_{ij}(t)$ and the observed data $Y_{t}$.}}
\label{F3}
\end{figure}
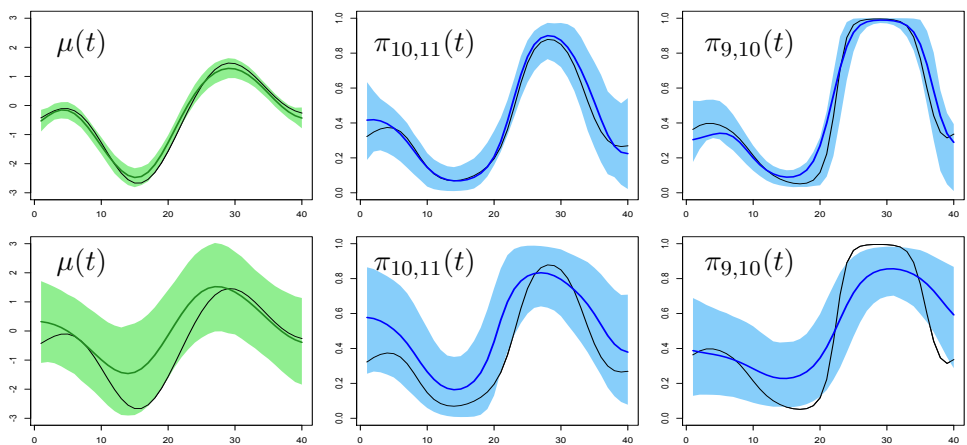
\begin{figure}[t]
\centering
\includegraphics[height=6cm, width=13cm]{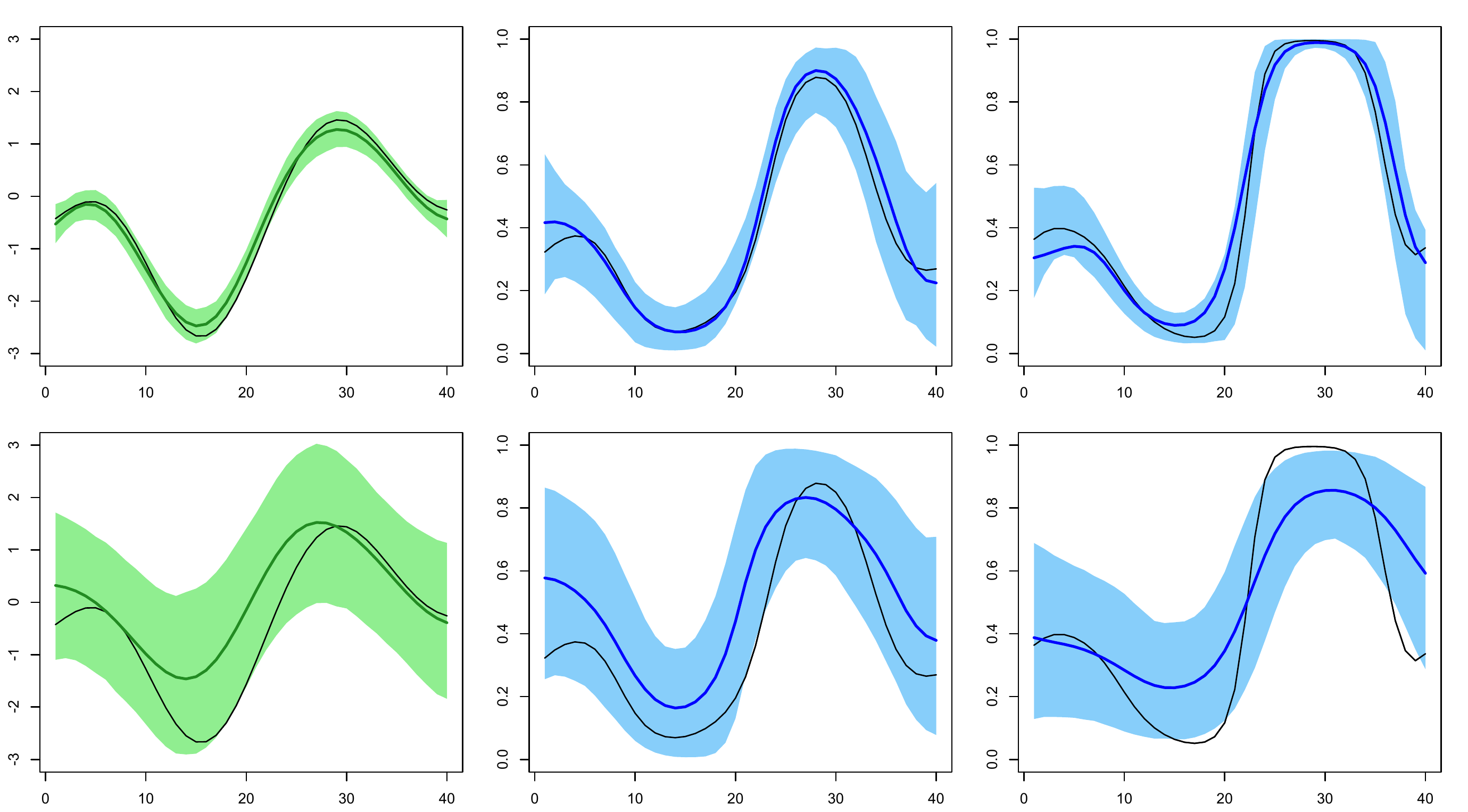}
\put (-350,150) {$\mu(t)$}
\put (-350,67) {$\mu(t)$}
\put (-230,150) {$\pi_{10,11}(t)$}
\put (-230,67) {$\pi_{10,11}(t)$}
\put (-105,150) {$\pi_{9,10}(t)$}
\put (-105,67) {$\pi_{9,10}(t)$}
\caption{\footnotesize{Top: For $\mu(t)$ and some selected $\pi_{ij}(t)$, plot of the true trajectory (black line), point-wise posterior mean (colored lines) and 0.95 highest posterior density (hpd) intervals (colored areas) for our model. Bottom: Same quantities estimated using only temporal information without exploring network structure (i.e. estimate $\pi_{ij}(t)$ using only the time series of the corresponding $y_{ij,t}$) }}
\label{F4}
\end{figure}
Figure \ref{F4} shows a graphical comparison between the performance of our model with respect to $\mu(t)$ and some selected probability trajectories $\pi_{ij}(t)$ (top), and the inferential results when the mean process and probability process $\pi_{ij}(t)$ are estimated with the same setting of our model but using only the time series of the corresponding $y_{ij,t}$ without borrowing information across the network (bottom). The sub-optimality of the independent approach is apparent in terms of both bias (over-smoothed trajectories) and variance (larger hpd intervals). When network structure is taken into account, the model provides accurate estimates, with posterior distributions rapidly concentrating around the true corresponding processes, while accurately selecting the dimension of the latent space.  In particular, we find that the estimated $\hat{\tau}_h^{-1}$ values start at 0.8 and 0.7 for $h=1$ and $2$, respectively, but then drop to small values for the later factors.  This implies that these later factor trajectories are quite flat and have limited influence.  Borrowing information across the network over time has the additional advantage of reducing hyperparameter sensitivity, in particular with respect to the length scale in GP prior.  We obtain, in fact, similar results when instead letting $\kappa_{\mu}=\kappa_{x}=0.03$, $\kappa_{\mu}=\kappa_{x}=0.1$ and $\kappa_{\mu}=\kappa_{x}=0.5$ in sensitivity analyses.

\section{Application to co-movements among National Stock Market Indices}
National Stock Indices represent technical tools constructed by a synthesis of numerous data on the evolution of the various stocks, and represent important indicators of the financial condition in a given country. Modeling co-variations among these quantities, and in general among assets, represents a fundamental issue in many financial applications, such as the Arbitrage Pricing Theory (APT) of \citet{Ross:1976} and the Capital Asset Pricing Model (CAPM) developed by \citet{Sha:1964}, and the correlations or covariances among asset's returns are the typical measures of co-movements employed in this framework. 

A rich literature is available in modeling dynamic covariance or correlation matrices, covering multivariate generalizations of ARCH and GARCH models (see e.g. \citealp{Tsay:2005}, \citealp{Eng:2002}, \citealp{Ale:2001}, \citealp{Boll:1988}), Stochastic volatility models \citep{Har:1994} and recent Bayesian extensions (see e.g. \citealp{Wi:2010}, \citealp{Naka:2012}, \citealp{Dur:2013}). In this application, we instead provide a different and not fully explored measure of co-movement exploiting the network structure among financial indices and giving exactly the probability that such event happens at a given time. This is accomplished by applying our model to the time-varying $Y_{t}$ matrices having entries $y_{ij,t}=y_{ji,t}=1$ if index $i$ and index $j$ co-move at time $t$ (indices are similar), and $y_{ij,t}=y_{ji,t}=0$ if opposite increments are recorded (indices are dissimilar). 
\begin{figure}[t]
\centering
\includegraphics[height=7cm, width=12cm]{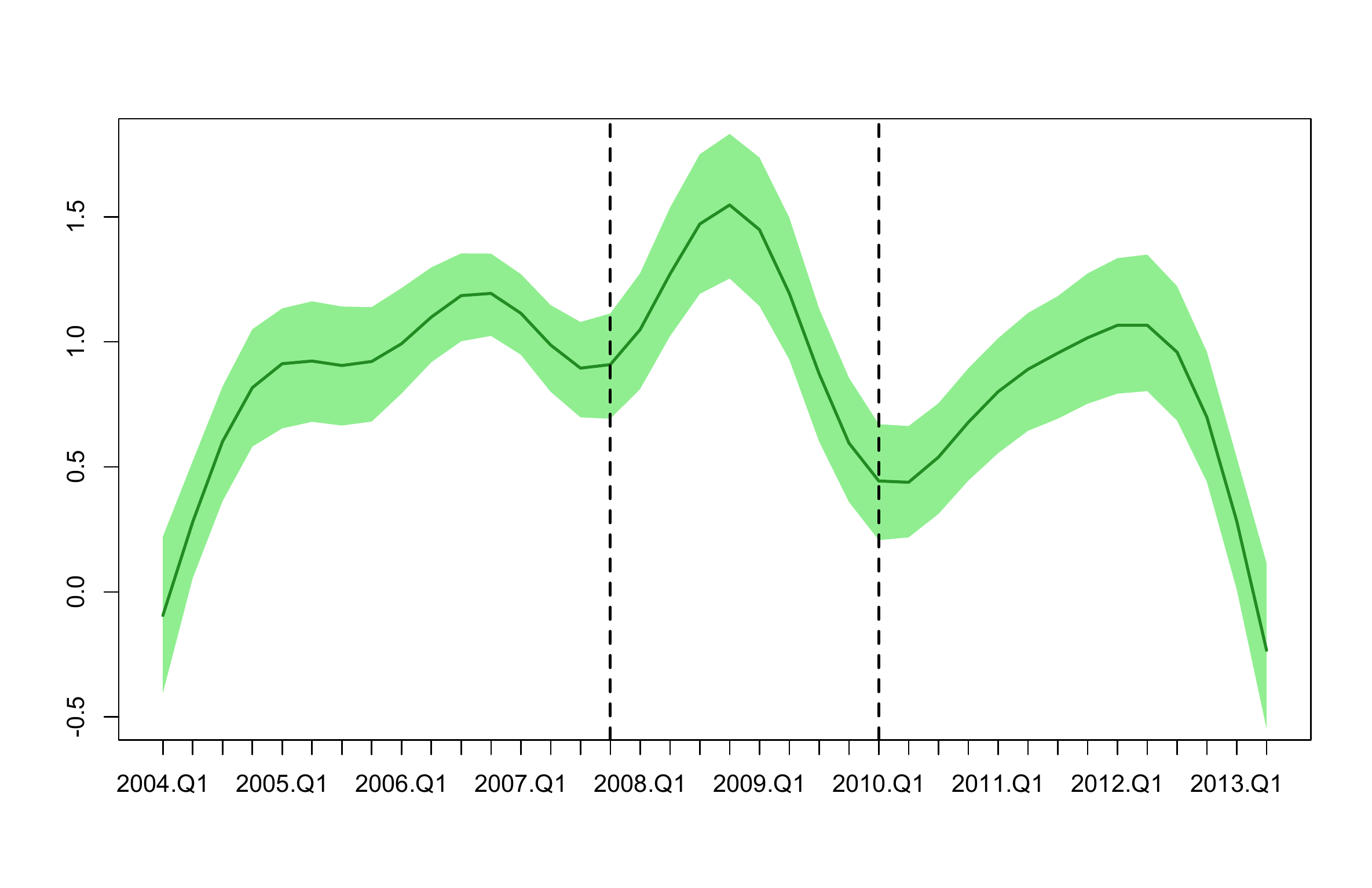}
\put (-255,55) {(A)}
\put (-165,55) {(B)}
\put (-80,55) {(C)}
\caption{\footnotesize{Plot of the point-wise posterior mean for the baseline $\mu(t)$ (colored line), and 0.95 hpd intervals (colored areas). (A) Growth and burst of USA housing bubble, (B) Global financial crisis, (C) Greek debt crisis, worsening of European sovereign-debt crisis and the rejection of the U.S. budget.}}
\label{F5}
\end{figure}

We constructed $Y_{t}$ using the quarterly log-returns of the $23$ main National Stock Market Indices ($V=23$) from 2004 to 2013 ($T=39$, with the last empty matrix $Y_{39}$ used for prediction), available at \url{ http://finance.yahoo.com/} and applied model (\ref{eq1}), with probabilities specified as in (\ref{eq2}) and latent similarity measures obtained via the projection approach defined in (\ref{eq3}). For posterior computation we run $5{,}000$ Gibbs iterations with a burn-in of $1{,}000$, setting a truncation level $H^{*}=15$, length scales $\kappa_{\mu}=0.03$, $\kappa_{x}=0.01$ and $a_1=a_2=2$. Similarly to the simulation study, most of the chains have effective sample sizes around $1{,}600$ on a total of $4{,}000$ after burn-in, showing good mixing.  We find that the first two latent factors are the most informative, with the remaining $13$ latent processes being concentrated near zero. A similar result was obtained in the seminal work of \citet{Fam:1993}, providing three main common risk factors in the returns of stocks.

\begin{figure}[t]
\centering
\includegraphics[height=4.5cm, width=15cm]{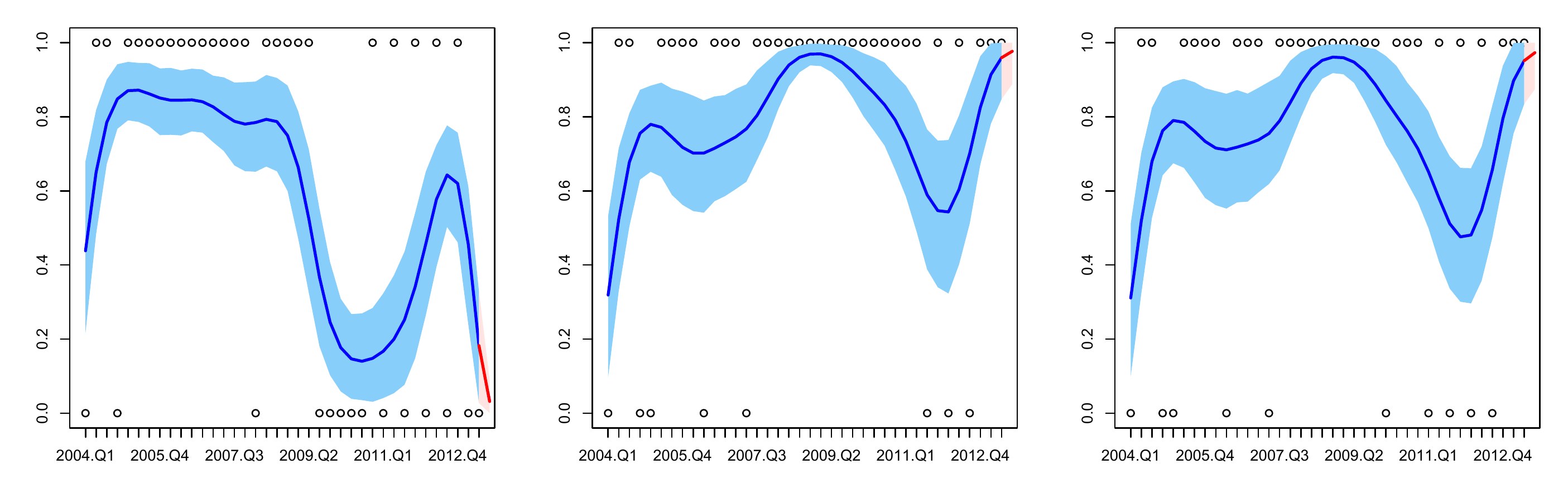}
\put (-385,125) {USA/Greece}
\put (-245,125) {USA/Germany}
\put (-100,125) {USA/France}
\caption{\footnotesize{Observed data (black dots), estimated co-movement probability trajectories (blue lines) and 0.95 highest posterior density intervals (colored blue areas), among USA and some selected European countries. Red lines and areas represent the same quantities with respect to posterior predictive distribution.}}
\label{F6}
\end{figure}

\subsection{Model Interpretation}
The estimated trajectory of the baseline process $\mu(t)$ together with the point-wise 0.95 hpd intervals in Figure \ref{F5}, provide important insights on the overall financial market behavior, in agreement with other theories on financial crises (see, e.g., \citealp{Bai:1999}, and \citealp{Cl:2009}) and recent applications (\citealp{Dur:2013}, \citealp{Kas:2013}). Increasing and persistent level of the baseline process, inducing higher probability of co-movements, are recorded during the growth and burst of USA housing bubble and the initial turmoils before the 2008 global financial crisis (A). This result provides an empirical proof in favor of the increasing inter-connection among financial markets due to the proliferation of risky loans between 2004 and 2007, and the growing demand by foreign countries for financial assets built from the real estate market, such as residential mortgage-backed securities (RMBS) and collateralized debt obligations (CDO). As expected the global financial crisis between late-2008 and end-2009 (B), and the following, Greek debt crisis together with the worsening of European sovereign-debt crisis (C), are manifested through a further increase of the co-movement probabilities, highlighting a clear financial contagion effect.

\begin{figure}[t]
\centering
\includegraphics[height=6cm, width=16cm]{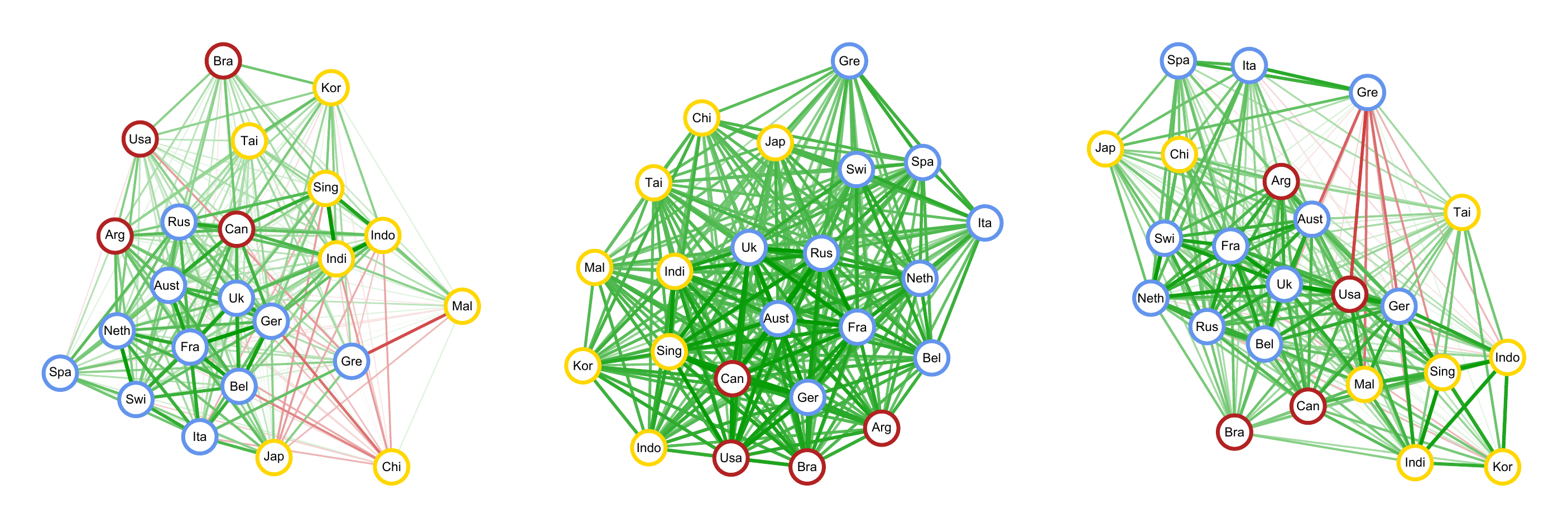}
\put (-450,170) {\small{(a) Aggregated Network}}
\put (-320,170) {\small{(b) World Financial Crisis Network}}
\put (-155,170) {\small{(c) Greek Debt Crisis Network}}
\caption{\footnotesize{Left: weighted network visualization with weights obtained averaging $\hat{\pi}(t)$ over $\mathcal{T}_0$. Middle: weighted network visualization with weights obtained averaging $\hat{\pi}(t)$ over the period of Global Financial Crisis (end 2008, beginning 2009). Right: Middle: weighted network visualization with weights obtained averaging $\hat{\pi}(t)$ over the period of Greek debts Crisis.  Edge dimensions are proportional to the corresponding value of the averaged probability matrix, with colors going from red to green gradation as the corresponding weight goes from $0$ to $1$. Blue, Red and Yellow nodes represent European, American and Asian countries, respectively.}}
\label{F7}
\end{figure}

Figure \ref{F6} shows the estimated (blue lines) and predicted (red lines) co-movement probability trajectories among USA and some selected European countries, pointing out the good performance of the proposed model in adaptively learning the data structure, confirmed also by a ROC curve having an area underneath of $0.79$. It is worth noticing that the local adaptivity of the estimated trajectories is not due to an over-parameterization of the model since the shrinkage prior on $\tau_h$ and the choice of small length scales in the GP covariance functions, imply smooth trajectories and a parsimonious model formulation. Thus adaptivity is provided by the information borrowed in the financial network for each time $t$. Co-movement probabilities among USA and Greece register a sharp drop in correspondence of the Greek debt crisis, differently from what happens with other European countries such as Germany and France, which instead evolve on similar patterns. We found this result reasonable in providing an empirical proof on the attempt to reduce the inter-connection with a country in crisis.

Finally, Figure \ref{F7} provides interesting insights on the financial network structure among the countries under investigation. Specifically we represent three different weighted networks, with weights given by the average estimated co-movement probability over all the time window considered (a), the estimated probability averaged over the period of the global financial crisis (b), and the Greek debt crisis (c). A reasonable global network structure with countries having similar financial economies most closely related among each other is provided in plot (a). As expected Japan appears to be closer to Western economies than Asian financial markets, while China has lower inter-connections with other countries. Stronger networks are estimated for European markets and Asian Tigers. International financial contagion effect is highlighted through strong inter-connections among all financial markets during the 2008 global financial crisis (b), with a still evident clustering effect, and Greece already showing a slightly different behavior. Finally, when the network during the Greek debt crisis is analyzed, we register evident low connections among Greece and almost all the other financial markets considered, and interestingly learn a strong network between Greece, Spain and Italy, representing the countries most affected by the European sovereign-debt crisis.

\section{Discussion}
We proposed a Bayesian nonparametric dynamic model for binary similarity matrices, borrowing information across time and the network structure of the data under investigation and allowing for dimensionality reduction. The model has been constructed using latent similarity measures defined by the dot product of latent coordinate vectors, with entries evolving in continuous time via Gaussian process priors. The shrinkage hyperprior allows us to automatically learn the dimension of the latent space and ensures a parsimonious definition of the model, with the risk of over-parameterization due to a higher number of latent features avoided. The P\'olya-Gamma data augmentation strategy allows us to define a simple and efficient Gibbs sampler for posterior computations based on full conditional conjugate posterior distributions, which is promising in terms of scaling to moderately large $V$, and easily handling missing values as well as forecasting problems. Scalability to large $T$ could be, instead, improved via stochastic differential equations models approximating the GP prior on the latent coordinate processes \citep{Zhu:2012}. We provided also theoretical results on the flexibility of the model, illustrated its performance via a simulation study and obtained interesting insights on the network among financial markets during the recent crisis, by applying the model to time-varying co-movement data.

Our model has a broad range of applicability, with dynamic social network analysis and time-varying binary evaluations among units providing two natural fields of application. Further directions of research could be devoted to the definition of similar models for discrete valued dynamic matrices, which could provide useful tools for analyzing  edge valued dynamic social networks or datasets with comparison among units expressed on a Likert scale.

\input{bibliografia}

\section{Appendix}
{\bf Proof of Theorem \ref{teo2}:} Since $\mathcal{T}$ is compact, for every $\epsilon_{0}>0$ there exists an open covering of $\epsilon_0$-balls $B_{\epsilon_0}(t_0):\{t: ||t-t_0 ||_2<\epsilon_0 \}$ with a finite subcover such that $\mathcal{T} \subset \cup_{t_0 \in \mathcal{T}_0}B_{\epsilon_0}(t_0)$, where $|\mathcal{T}_0|=T$. Then:
$$ \Pi_{S} \left( \sup_{t \in \mathcal{T}}||S(t)-S^*(t) ||_2<\epsilon \right)= \Pi_{S} \left(\max_{t_0 \in \mathcal{T}_0} \sup_{t \in B_{\epsilon_0}(t_0)}||S(t)-S^*(t) ||_2<\epsilon \right).$$
Define $Z(t_0)=\sup_{t \in B_{\epsilon_0}(t_0)}||S(t)-S^*(t) ||_2$. Since
$$\Pi_{S} \left(\max_{t_0 \in \mathcal{T}_0} Z(t_0)<\epsilon \right)>0 \Longleftrightarrow \Pi_{S} \left(Z(t_0)<\epsilon \right) >0, \ \forall t_0 \in \mathcal{T}_0$$
we only need to look at each $\epsilon_0$-ball independently as follow:
{\footnotesize{
\begin{eqnarray}
\label{eq9}
&&\Pi_{S} \left( \sup_{t \in B_{\epsilon_0}(t_0)}||S(t)-S^*(t) ||_2<\epsilon\right) \\
 &\geq& \Pi_{S} \left(||S(t_0)-S^*(t_0) ||_2+\sup_{t \in B_{\epsilon_0}(t_0)}||S^*(t_0)-S^*(t) ||_2+ \sup_{t \in B_{\epsilon_0}(t_0)}||S(t_0)-S(t) ||_2<\epsilon\right) \nonumber\\
&\geq& \Pi_{S} \left(||S(t_0)-S^*(t_0) ||_2<\frac{\epsilon}{3}\right)  \Pi_{S}\left(\sup_{t \in B_{\epsilon_0}(t_0)}||S^*(t_0)-S^*(t) ||_2<\frac{\epsilon}{3}\right)\Pi_{S}\left( \sup_{t \in B_{\epsilon_0}(t_0)}||S(t_0)-S(t) ||_2<\frac{\epsilon}{3}\right) \nonumber
\end{eqnarray}}}
Where the first inequality comes from repeated uses of triangle inequality, and the second follows from the fact that each of these terms is an independent event. We evaluate each of these terms in turn.

Based on the continuity of $S^{*}(\cdot)$, for all $\epsilon/3>0$, there exists an $\epsilon_{0,1}>0$ such that:
$$||S(t_0)-S^*(t_0) ||_2<\frac{\epsilon}{3}, \quad \forall||t-t_0 ||_2<\epsilon_{0,1} $$
Therefore, $\Pi_{S}\left(\sup_{t \in B_{\epsilon_{0,1}}(t_0)}||S^*(t_0)-S^*(t) ||_2<\frac{\epsilon}{3}\right)=1$.

Given the GP prior on the elements of $X(\cdot)$ and letting $x_{ih}(t)=[X(t)]_{ih}$, the equation
$$[X(t)X(t)']_{ij}=\sum_{h=1}^{H}x_{ih}(t)x_{jh}(t), \quad \forall t \in \mathcal{T} $$ 
represents a finite sum over pairwise products of almost surely continuous functions (recalling GP assumption on the elements $x_{ih}$) and thus result in a matrix $X(t)X(t)'$ with elements almost surely continuous on $\mathcal{T}$. Therefore $S(t)=\mu(t)\times 1_V1_V'+X(t)X(t)'$ is almost surely continuous on $\mathcal{T}$ since the baseline $\mu(\cdot)$ is itself almost surely continuous given the GP prior assumption. Therefore, similarly as before, for all $\epsilon/3>0$, there exists and $\epsilon_{0,2}>0$ such that:
$$\Pi_{S}\left(\sup_{t \in B_{\epsilon_{0,2}}(t_0)}||S(t_0)-S(t) ||_2<\frac{\epsilon}{3}\right)=1$$

To examine last term, first note that:
{\footnotesize{
$$\Pi_{S}\left(||S(t_0)-S^{*}(t_0) ||_2 <\frac{\epsilon}{3} \right)=\Pi_{S} \left(||\mu(t_0)\times 1_V1_V'+X(t_0)X(t_0)' -\mu^*(t_0)\times 1_V1_V'-X(t_0)^*X(t_0)^{*'}||_2<\frac{\epsilon}{3}\right)$$}}
Where $\{ X(t_0)^*,\mu^*(t_0) \}$ is any element of $\mathcal{X}_{X} \otimes \mathcal{X}_{\mu}$ such that $S^{*}(t_0)=\mu^*(t_0)\times 1_V1_V'+X(t_0)^*X(t_0)^{*'}$. Such a factorization exists by Corollary 2. Thus, using triangle inequality, we can bound this probability by:
\begin{eqnarray*}
&&\Pi_{S}\left(||S(t_0)-S^{*}(t_0) ||_2 <\frac{\epsilon}{3} \right)\\
&\geq& \Pi_{S} \left(||X(t_0)X(t_0)'-X(t_0)^*X(t_0)^{*'}||_2<\frac{\epsilon}{6}\right)\Pi_{\mu}\left(|| 1_V1_V'( \mu(t_0)-\mu^*(t_0)||_2<\frac{\epsilon}{6}\right)
\end{eqnarray*}
Based on the support of the Gaussian prior,
$$\Pi_{\mu}\left(|| 1_V1_V'( \mu(t_0)-\mu^*(t_0)||_2<\frac{\epsilon}{6}\right)=\Pi_{\mu}\left(|( \mu(t_0)-\mu^*(t_0)|<\frac{\epsilon}{6\sqrt{V}}\right)>0.$$
For studying the first term of the previous decomposition note that:
$$X(t_0)X(t_0)'| \{\tau_{h} \}_{h=1}^{H} = \sum_{h=1}^{H} x_h(t_0) x_h(t_0)' $$
where $x_h(t_0)=[x_{1h}(t_0), \dots x_{Vh}(t_0)]'$ is distributed, according to our prior specification, as $\mbox{N}_V(0,\tau_{h}^{-1}I_{V})$, implying that $x_h(t_0) x_h(t_0)'|\tau_{h} \sim \mbox{W}_{V}(1,\tau_{h}^{-1}I_{V})$ independently for all $h=1,...,H$, where $\mbox{W}_{V}(\cdot, \cdot)$ denotes the Wishart random variable. Using the triangle inequality we obtain:
{\footnotesize{
$$ \Pi_{S} \left(||X(t_0)X(t_0)'-X(t_0)^*X(t_0)^{*'}||_2<\frac{\epsilon}{6}\right)
\geq \prod_{h=1}^{H} \Pi_{x_h}\left(||x_h(t_0)x_h(t_0)'-x_h(t_0)^*x_h(t_0)^{*'}||_2<\frac{\epsilon}{6H}\right)$$}}
Since $x_h(t_0)^*x_h(t_0)^{*'}$ is an arbitrary rank-1 symmetric matrix in $\Re^{V \times V}$, and based on the support of the Wishart distribution:
$$\Pi_{x_h}\left(||x_h(t_0)x_h(t_0)'-x_h(t_0)^*x_h(t_0)^{*'}||_2<\frac{\epsilon}{6H}\right)>0, \quad \forall h=1,...,H. $$
Thus $\Pi_{S} \left(||X(t_0)X(t_0)'-X(t_0)^*X(t_0)^{*'}||_2<\frac{\epsilon}{6}\right)>0$ and combining it with the large support property previously proved for the prior on the baseline $\mu(\cdot)$, we have:
$$\Pi_{S}\left(||S(t_0)-S^{*}(t_0) ||_2 <\frac{\epsilon}{3} \right)>0$$
For every $S^{*}(\cdot)$ and $\epsilon>0$, let $\epsilon_{0}=\min(\epsilon_{0,1},\epsilon_{0,2})$, with $\epsilon_{0,1}$ and $\epsilon_{0,2}$ defined as above. Then, combining the positivity results of each of the three terms in \ref{eq9} completes the proof.

\end{document}

%% file: bibliografia.tex

%% file: DSM.bbl
\begin{thebibliography}{}
\addcontentsline{toc}{chapter}{Bibliography}


\bibitem[\protect\citeauthoryear{Airoldi et al.}{Airoldi et al.}{2008}]{Ai:2008}
Airoldi, E.M., Blei, D.M., Fienberg, S.E., \& Xing, E.P. (2008).
\newblock Mixed Membership Stochastic Blockmodels.
\newblock {\em Journal of Machine Learning Research\/}~{\em 9\/}, 1981--2014.

\bibitem[\protect\citeauthoryear{Alexander}{Alexander}{2001}]{Ale:2001}
Alexander, C.O. (2001).
\newblock Orthogonal GARCH.
\newblock {\em Mastering Risk\/}~{\em 2\/}, 21--38.

\bibitem[\protect\citeauthoryear{Baig and Goldfaijn}{Baig and Goldfaijn}{1999}]{Bai:1999}
Baig, T., \& Goldfaijn, I. (1999).
\newblock Financial Market Contagion in the Asian Crisis.
\newblock {\em Staff Papers, International Monetary Fund\/}~{\em 46\/}, 167--195.




\bibitem[\protect\citeauthoryear{Bhattacharya \& Dunson}{Bhattacharya \& Dunson}{2011}]{Bhat:2011}
Bhattacharya, A., \& Dunson, D.B. (2011).
\newblock Sparse Bayesian infinite factor models.
\newblock {\em Biometrika\/}~{\em 98\/}, 291--306.

\bibitem[\protect\citeauthoryear{Bollen}{Bollen}{1989}]{Bo:1989}
Bollen, K.A. (1989).
\newblock {\em Structural Equations with Latent Variables.}
\newblock Wiley.

\bibitem[\protect\citeauthoryear{Bollerslev et al.}{Bollerslev et al.}{1988}]{Boll:1988}
Bollerslev, T., Engle, R.F., \& Wooldridge, J.M. (1988).
\newblock A capital-asset pricing model with time-varying covariances.
\newblock {\em Journal of Political Economy\/}~{\em 96\/}, 116--131.

\bibitem[\protect\citeauthoryear{Claessens \& Forbes}{Claessens \& Forbes}{2009}]{Cl:2009}
Claessens, S., \& Forbes, K. (2009)
\newblock International Financial Contagion, An overview of the Issues.
\newblock Springer.


\bibitem[\protect\citeauthoryear{Cox \& Cox}{Cox \& Cox}{2001}]{Cox:2001}
Cox, T.F., \& Cox, M.A. (2001).
\newblock {\em Multidimensional Scaling.}
\newblock II ed., Chapman and Hall.


\bibitem[\protect\citeauthoryear{DeSarbo et al.}{DeSarbo et al.}{1999}]{De:1999}
DeSarbo, W.S., Kim, Y., \& Fong D. (1999).
\newblock  A Bayesian multidimensional scaling procedure for the spatial analysis of revealed choice data.
\newblock {\em Journal of Econometrics\/}~{\em 89\/}, 79--108.

\bibitem[\protect\citeauthoryear{DeSarbo \& Hoffman}{DeSarbo \& Hoffman}{1987}]{De:1987}
DeSarbo, W.S., \& Hoffman, D.L. (1987).
\newblock  Constructing MDS joint spaces from binary choice data: A new multidimensional unfolding threshold model for marketing research.
\newblock {\em Journal of Marketing
Research\/}~{\em 24\/}, 40--54.

\bibitem[\protect\citeauthoryear{Durante et al.}{Durante et al.}{2013}]{Dur:2013}
Durante, D., Scarpa, B. \& Dunson, D.B. (2013).
\newblock Locally adaptive factor processes for multivariate time series.
\newblock \url{http://arxiv.org/abs/1210.2022}.

\bibitem[\protect\citeauthoryear{Engle}{Engle}{2002}]{Eng:2002}
Engle, R.F. (2002).
\newblock Dynamic conditional correlation: a simple class of multivariate generalized autoregressive conditional heteroskedasticity models.
\newblock {\em Journal of Business \& Economic Statistics\/}~{\em 20\/}, 339--350.


\bibitem[\protect\citeauthoryear{Erd\"{o}s \& R\'eny}{Erd\"{o}s \& R\'eny}{1959}]{Erd:1959}
Erd\"{o}s, P., \& R\'eny, A. (1959).
\newblock  On Random Graphs
\newblock {\em Publicationes Mathematicae\/}~{\em 6\/}, 290--297.


\bibitem[\protect\citeauthoryear{Fama \& French}{Fama \& French}{1993}]{Fam:1993}
Fama, E.F., \& French, K.R. (1993).
\newblock Common risk factors in the returns on stocks and bonds.
\newblock {\em Journal of Financial Economics\/}~{\em 33\/}, 3--56.



\bibitem[\protect\citeauthoryear{Frank \& Strauss}{Frank \& Strauss}{1986}]{Fra:1986}
Frank, O., \& Strauss, D. (1986).
\newblock Markov Graphs.
\newblock {\em Journal of the American Statistical Association\/}~{\em 81\/}, 832--842.


\bibitem[\protect\citeauthoryear{Ghosh \& Dunson}{Ghosh \& Dunson}{2009}]{Gosh:2009}
Ghosh, J., \& Dunson, D.B. (2009)
\newblock Default priors and efficient posterior computation in Bayesian factor analysis.
\newblock {\em Journal of Computational and Graphical Statistics\/}~{\em 18\/}, 306--320.



\bibitem[\protect\citeauthoryear{Guo et al.}{Guo et al.}{2007}]{Guo:2007}
Guo, F., Hanneke S., Fu, W., \& Xing, E.P. (2007).
\newblock Recovering temporally rewiring networks: A model-based approach.
\newblock {\em In International Conference in Machine Learning}.

\bibitem[\protect\citeauthoryear{Handcock et al.}{Handcock et al.}{2003}]{Han:2003}
Handcock, M.S., Robins, G.L., Snijders, T.A.B., Moody, J., \& Besag, J. (2003).
\newblock Assessing Degeneracy in Statistical Models of Social Networks.
\newblock {\em Journal of the American Statistical Association\/}~{\em 76\/}, 33--50.

\bibitem[\protect\citeauthoryear{Harvey et al.}{Harvey et al.}{1994}]{Har:1994}
 Harvey, A.C., Ruiz E., \& Shepard N. (1994).
\newblock Multivariate stochastic variance models.
\newblock {\em Review of Economic Studies,\/}~{\em 61\/}, 247--264.



\bibitem[\protect\citeauthoryear{Hoff et al.}{Hoff et al.}{2002}]{Hoff:2002}
Hoff, P.D., Raftery, A.E., \& Handcock, M.S. (2002).
\newblock Latent Space Approaches to Social Network Analysis.
\newblock {\em Journal of the American Statistical Association\/}~{\em 97\/}, 1090--1098.



\bibitem[\protect\citeauthoryear{Holbrook et al.}{Holbrook et al.}{1982}]{Hol:1982}
Holbrook M.B., William L.M., \& Russell S.W. (1982).
\newblock  Constructing Joint Spaces from "Pick-Any" Data: A New Tool for Consumer Analysis.
\newblock {\em Journal of Consumer Research\/}~{\em 9\/}, 99--105.

\bibitem[\protect\citeauthoryear{Holland \& Leinhardt}{Holland \& Leinhardt}{1981}]{Hol:1981}
Holland, P.W., \& Leinhardt, S. (1981).
\newblock An Exponential Family of Probability Distributions forDirected Graphs.
\newblock {\em Journal of the American Statistical Association\/}~{\em 76\/}, 33--65.

\bibitem[\protect\citeauthoryear{Ishiguro et al.}{Ishiguro et al.}{2010}]{Ish:2010}
Ishiguro, K., Iwata, T., Ueda, N. \& Tenenbaum, J. (2010).
\newblock Dynamic infinite relational model for time-varying relational data analysis.
\newblock {\em In Advances in Neural Information Processing Systems (NIPS)}.


\bibitem[\protect\citeauthoryear{Jamali-Rad \& Leus}{Jamali-Rad \& Leus}{2012}]{Jam:2012}
Jamali-Rad, H., \& Leus, G. (2012).
\newblock Dynamic Multidimensional Scaling for Low-Complexity Mobile Network Tracking.
\newblock {\em IEEE Transactions on Signal Processing \/}~{\em 60\/}, 4485--4491.

\bibitem[\protect\citeauthoryear{Kastner et al.}{Kastner et al.}{2013}]{Kas:2013}
Kastner, G., Fr\"{u}hwirth-Schnatter, S., \& Lopes, H.F. (2013)
\newblock Analysis of Exchange Rates via Multivariate Bayesian Factor Stochastic Volatility Models.
\newblock In Lanzarone E., \& Ieva, F. {\em The Contribution of Young Researchers to Bayesian Statistics, Proceedings of BAYSM2013.\/}~{\em 63\/}, Springer.


\bibitem[\protect\citeauthoryear{Kemp et al.}{Kemp et al.}{2006}]{Kem:2006}
Kemp, C., Tenenbaum, J. B., Griffiths, T. L., Yamada, T. \& Ueda, N. (2006).
\newblock Learning systems of concepts with an infinite relational model.
\newblock {\em In Proceedings of the 21st National Conference on Artificial Intelligence}.



\bibitem[\protect\citeauthoryear{Nakajima \& West}{Nakajima \& West}{2012}]{Naka:2012}
Nakajima, J., \& West, M. (2012)
\newblock Dynamic factor volatility modeling: A Bayesian latent threshold approach.
\newblock {\em Journal of Financial Econometrics\/}~{\em 11\/}, 116--153.

\bibitem[\protect\citeauthoryear{Nowicki \& Snijders}{Nowicki \& Snijders}{2001}]{Now:2001}
Nowicki, K., \& Snijders, T.A.B. (2001).
\newblock Estimation and prediction for stochastic blockstructures. 
\newblock {\em Journal of the American Statistical Association\/}~{\em 96\/}, 1077--1087.



\bibitem[\protect\citeauthoryear{Oh \& Raftery}{Oh \& Raftery}{2007}]{Oh:2007}
Oh, M.S., \& Raftery A.E. (2007).
\newblock  Model-based clustering with dissimilarities: A Bayesian approach.
\newblock {\em Journal of Computational and Graphical Statistics\/}~{\em 16\/}, 559--585.

\bibitem[\protect\citeauthoryear{Oh \& Raftery}{Oh \& Raftery}{2001}]{Oh:2001}
Oh, M.S., \& Raftery A.E. (2001).
\newblock Bayesian Multimensional scaling and choice of dimension.
\newblock {\em Journal of the American Statistical Association\/}~{\em 96\/}, 1031--1004.



\bibitem[\protect\citeauthoryear{Polson et al.}{Polson et al.}{2013}]{Po:2013}
Polson, N.G., Scott J.G., \& Windle J. (2013).
\newblock Bayesian inference for logistic models using Polya-Gamma latent variables.
\newblock \url{http://arxiv.org/abs/1205.0310}.

\bibitem[\protect\citeauthoryear{Robins \& Pattinson}{Robins \& Pattinson}{2001}]{Rob:2001}
Robins, G.L., \& Pattison, P.E., (2001).
\newblock Random graph models for temporal processes in social networks.
\newblock {\em Journal of Mathematical Sociology\/}~{\em 25\/}, 5--41.

\bibitem[\protect\citeauthoryear{Ross}{Ross}{1976}]{Ross:1976}
Ross, S. (1976).
\newblock The arbitrage theory of capital asset pricing.
\newblock {\em Journal of Finance\/}~{\em 13\/}, 341--360.


\bibitem[\protect\citeauthoryear{Sarkar et al.}{Sarkar et al.}{2007}]{Sark:2007}
Sarkar, P., Siddiqi, S.M., \& Gordon, G.J. (2007).
\newblock  A latent space approach to dynamic embedding of co-occurrence data.
\newblock {\em In Proceedings of the 11th International Conference on Artificial Intelligence and Statistics (AI-STATS).}

\bibitem[\protect\citeauthoryear{Sarkar \& Moore}{Sarkar \& Moore}{2005}]{Sark:2005}
Sarkar, P., \& Moore, A.W. (2005).
\newblock Dynamic social network analysis using latent space models.
\newblock {\em In Advances in Neural Information Processing Systems (NIPS)}.

\bibitem[\protect\citeauthoryear{Sharpe}{Sharpe}{1964}]{Sha:1964}
Sharpe, W. (1964).
\newblock Capital asset prices: a theory of market equilibrium under conditions of risk.
\newblock {\em Journal of Finance\/}~{\em 19\/}, 425--442.

\bibitem[\protect\citeauthoryear{Snijders}{Snijders}{2002}]{Sni:2002}
Snijders, T.A.B. (2002).
\newblock Markov Chain Monte Carlo Estimation of Exponential Random Graph Models.
\newblock {\em Journal of Social Structure\/}~{\em 2\/}, 1--40.




\bibitem[\protect\citeauthoryear{Strauss \& Ikeda}{Strauss \& Ikeda}{1990}]{Stra:1999}
Strauss, D., \& Ikeda, M. (1990).
\newblock Pseudolikelihood Estimation for Social Networks.
\newblock {\em Journal of the American Statistical Association\/}~{\em 49\/}, 204--212.

\bibitem[\protect\citeauthoryear{Tsay}{Tsay}{2005}]{Tsay:2005}
Tsay, R.S. (2005).
\newblock {\em Analysis of Financial Time Series.}
\newblock II ed., Wiley.

\bibitem[\protect\citeauthoryear{Wilson \& Ghahramani}{Wilson \& Ghahramani}{2010}]{Wi:2010}
Wilson, A.G. \& Ghahramani Z. (2010).
\newblock Generalised Wishart Processes.
\newblock \url{http://arxiv.org/abs/1101.0240}.



\bibitem[\protect\citeauthoryear{Xing et al.}{Xing et al.}{2010}]{Xi:2010}
Xing, E.P., Fu, W., \& Song, L. (2010).
\newblock A State-Space Mixed Membership Blockmodel for Dynamic Network Tomography.
\newblock {\em The Annals of Applied Statistics\/}~{\em 4\/}, 535--566.


\bibitem[\protect\citeauthoryear{Xu \& Hero}{Xu \& Hero}{2013}]{Xu:2013}
Xu, S.K., \& Hero III, A.O., (2013).
\newblock Dynamic stochastic blockmodels: Statistical models for time-evolving networks
\newblock \url{http://arxiv.org/abs/1304.5974}.

\bibitem[\protect\citeauthoryear{Yang et al.}{Yang et al.}{2011}]{Ya:2011}
Yang, T.B, Chi, Y., Zhu, S.H., Gong, Y.H., \& Jin, R. (2011).
\newblock Detecting communities and their evolutions in dynamic social networks--a
Bayesian approach.
\newblock {\em Machine Learning\/}~{\em 82\/}, 157--189.

 
\bibitem[\protect\citeauthoryear{Zhu and Dunson}{Zhu and Dunson}{2012}]{Zhu:2012}
Zhu, B., \& Dunson, D.B. (2012).
\newblock Locally Adaptive Bayes Nonparametric Regression via Nested Gaussian Processes.
\newblock \url{http://arxiv.org/abs/1201.4403}.
\end{thebibliography}
